\documentclass{article}
\usepackage{spconf,amsmath,graphicx}
\usepackage{booktabs}
\usepackage{subcaption}
\usepackage{adjustbox}
\usepackage{amssymb}

\usepackage{enumitem}
\setlist{nosep, leftmargin=14pt}
\usepackage{mwe} 

\title{Rotation Equivariant Convolutions in Deformable Registration of Brain MRI}
\name{Arghavan Rezvani, Kun Han, Anthony T. Wu, Pooya Khosravi, Xiaohui Xie}
\address{University of California, Irvine}
\begin{document}
%
\maketitle
\begin{abstract}
Image registration is a fundamental task that aligns anatomical structures between images. While CNNs perform well, they lack rotation equivariance -- a rotated input does not produce a correspondingly rotated output. This hinders performance by failing to exploit the rotational symmetries inherent in anatomical structures, particularly in brain MRI.
In this work, we integrate rotation-equivariant convolutions into deformable brain MRI registration networks. We evaluate this approach by replacing standard encoders with equivariant ones in three baseline architectures, testing on multiple public brain MRI datasets.

Our experiments demonstrate that equivariant encoders have three key advantages: 1) They achieve higher registration accuracy while reducing network parameters, confirming the benefit of this anatomical inductive bias. 2) They outperform baselines on rotated input pairs, demonstrating robustness to orientation variations common in clinical practice. 3) they show improved performance with less training data, indicating greater sample efficiency. Our results demonstrate that incorporating geometric priors is a critical step toward building more robust, accurate, and efficient registration models.
\end{abstract}







%
%

\section{Introduction}
\label{sec:intro}
Deformable image registration, which aligns anatomical structures between image pairs, is a fundamental task in medical image analysis. Traditional optimization-based methods achieve accurate alignment but are computationally expensive. Learning-based methods, popularized by VoxelMorph \cite{balakrishnan2019voxelmorph}, revolutionized the field by using CNNs to predict deformation fields directly, enabling fast and accurate unsupervised alignment. Subsequent works have improved this paradigm, such as Dual-PRNet++ \cite{kang2022dual} which incorporates explicit voxel correspondences, and RDP Net \cite{wang2024recursive} which employs recursive refinement.
However, standard CNNs are only translation-equivariant. This design is brittle to a core clinical reality: unavoidable variations in patient positioning (e.g., due to patient discomfort or height) that introduce rotational shifts. These networks, lacking rotational equivariant features, thus generalize poorly to novel orientations. Furthermore, they fail to exploit the local rotational symmetry of anatomical features, such as the gyri and sulci in brain MRI, leading to inefficient and suboptimal feature representations.

Recent advances in geometric deep learning offer a solution. While Group-CNNs \cite{cohen2016group} achieve equivariance to finite groups at a high memory cost, Steerable CNNs \cite{cohen2016steerable} efficiently handle continuous groups, like SE(3) in 3D, by imposing mathematical constraints on kernels. In medical imaging, SE(3)-equivariant kernels have already improved segmentation \cite{diaz2024leveraging}, rigid motion tracking \cite{billot2024se}, and affine registration \cite{wang2024rotir}, demonstrating clear benefits in robustness.

In this work, we extend SE(3) equivariant kernels to deformable registration, a direction that remains unexplored. Specifically, we replace the standard convolutional encoders in three representative architectures—VoxelMorph, Dual-PRNet++, and RDP Net—with SE(3)-equivariant encoders. Our \textbf{contributions} are four-fold: We demonstrate that equivariant encoders (1) improve registration accuracy on brain MRI with fewer parameters; (2) show improved robustness to orientation variations; (3) exhibit greater sample efficiency with reduced training data; and (4) We investigate the effect of steerable kernel parameters on performance.

\section{Method}
\label{sec:method}
\textbf{2.1. Problem Formulation:}
Given a moving and fixed image $I_m , I_f \in \mathbb{R}^{H \times W \times D}$, registration aims to find a dense deformation field $\phi: \Omega \to \Omega$ (where $\Omega \subset \mathbb{R}^3$) to align the anatomical structures of the warped moving image $I_m\circ \phi $ with the fixed image $I_f$.
This process minimizes a loss function of the form: 
\vspace{-15pt}
\begin{equation}
    \mathcal{L}(\phi) = \mathcal{L}_{\text{sim}}(I_f, I_m \circ \phi) + \lambda \mathcal{L}_{\text{reg}}(\phi)
    \vspace{-8pt}
\end{equation}
 where $L_{sim}$ measures the similarity between the aligned images, $L_{reg}$ enforces spatial smoothness of the deformation field and $\lambda$ is a regularization weight. In this work, $\mathcal{L}_{\text{sim}}$ is a local normalized cross-correlation (NCC) loss, and $\mathcal{L}_{\text{reg}}$ is the $\ell_2$-norm of the spatial gradient of the displacement field.

\noindent
\textbf{2.2. Baseline Architectures:}
To evaluate the impact of rotation equivariance on deformable registration, we selected \textit{three representative baseline architectures}. We replaced their standard convolutional encoders with rotation-equivariant encoders—keeping the decoders unchanged—to isolate the effect of equivariant feature extraction: 
\textbf{VoxelMorph (VM):} A single-stage U-Net that takes concatenated image pairs and directly predicts dense deformation fields.
\textbf{Dual-PRNet++:} A multi-stage architecture with dual-stream encoders  processing images separately, incorporating pyramid registration with 3D correlation volumes.
\textbf{RDP Net:} A recursive pyramid network employing coarse-to-fine registration with multiple resolution levels and recursive refinement.

These architectures were chosen for their diversity, covering key design dimensions: single vs. multi-stage processing, joint vs. separate input encoding, and complexities ranging from a simple U-Net to pyramid-based correlation and recursive refinement. This diversity ensures our findings represent a general principle applicable across various registration frameworks. 

Henceforth, we refer to the equivariant-encoder variants as Equi-VM, Equi-Dual-PRNet++, and Equi-RDP Net.




\noindent
\textbf{2.3. Steerable Kernels:}
Standard convolutions operate on scalar-valued features, limiting their ability to capture geometric relationships fundamental to medical image registration. To enforce consistent geometric behavior under rotations and translations, we adopt \textit{equivariant} mappings constructed with steerable kernels.

A function $\Phi : \mathcal{F}_{\text{in}} \to \mathcal{F}_{\text{out}}$ between feature fields is \textit{equivariant} to a transformation group $G$ if, for all $g \in G$,
\vspace{-8pt}
\begin{equation}
  \Phi(T_g f) = T'_g(\Phi(f)),
\vspace{-6pt}
\end{equation}
where $T_g$ and $T'_g$ denote the actions of $G$ on the input and output spaces, respectively. 

For the $SE(3)$ group, which combines rotations $R \in SO(3)$ and translations $\mathbf{t} \in \mathbb{R}^3$, the group action on a feature field $f : \mathbb{R}^3 \to \mathbb{R}^C$ is defined as
\vspace{-8pt}
\begin{equation}
   [T_{(R,\mathbf{t})} f](\mathbf{x}) = \rho(R)\, f(R^{-1}(\mathbf{x} - \mathbf{t})),
\vspace{-6pt}
\end{equation}
where $\rho(R)$ is a finite-dimensional representation of $SO(3)$ describing how the feature channels transform under rotation. These channel transformations can represent scalars, vectors, or higher-order geometric features, providing richer geometric structure than standard CNNs.

Any finite-dimensional representation of $SO(3)$ can be decomposed into irreducible representations (irreps) indexed by $l = 0, 1, 2, \dots$, with dimension $d_l = 2l + 1$. We restrict our analysis to three fundamental irreps: scalars ($l=0$), vectors ($l=1$), and second-rank tensors ($l=2$), balancing geometric expressiveness with computational efficiency for 3D medical image registration.

Following Weiler \emph{et al.}~\cite{weiler20183d}, steerable 3D convolution
kernels are constructed from a basis that separates \emph{radial} and
\emph{angular} dependencies:
\vspace{-8pt}
\begin{equation}
    \kappa(\mathbf{x})
    = \sum_{l} \Phi_l(\|\mathbf{x}\|)\, Q_l(\mathbf{x}/\|\mathbf{x}\|),
    \vspace{-4pt}
\end{equation}
where $\Phi_l(r)$ are learnable radial functions, $Q_l(\mathbf{x}/\|\mathbf{x}\|)$ are angular basis elements derived from spherical harmonics.
This parameterization ensures that the kernel satisfies the equivariance constraint:
\vspace{-6pt}
\begin{equation}
    \rho^{(l_{\text{out}})}(R)\, \kappa(\mathbf{x})\, \rho^{(l_{\text{in}})}(R)^{-1} 
    = \kappa(R\mathbf{x}), \quad \forall R \in SO(3),
    \vspace{-3pt}
\end{equation}
guaranteeing $SO(3)$-equivariance by construction. Since convolution is translation-equivariant, the resulting layer is fully $SE(3)$-equivariant.


\noindent
\textbf{2.4. Equivariant Encoder Design:}
\label{sec:enc_design}
For each baseline, we design an SE(3)-equivariant encoder maintaining equivariance to 3D rotations and translations, replacing the standard convolutional encoder while preserving the original decoder to isolate the impact of equivariant feature extraction on registration performance.

Each equivariant block consists of SE(3)-steerable convolution followed by a gated activation function \cite{weiler20183d}:
For each non-scalar irreducible feature $f_n^{(l)}(\mathbf{x})$ of irrep-$l$ in layer $n$, we generate a scalar gate using a separate steerable convolution $g(\mathbf{x}) = \gamma\ast f_{n-1}(\mathbf{x})$, apply sigmoid activation $s(\mathbf{x}) = \sigma(g(\mathbf{x}))$, and multiply: $\tilde{f}_n^{(l)}(\mathbf{x}) = s(\mathbf{x}) \cdot f_n^{(l)}(\mathbf{x})$, where $\gamma$ is a learnable SE(3)-steerable kernel producing scalar output.
The architectural parameters (number of convolutional kernels used at each level, stride, padding, and kernel size) match baselines for fair comparison. 
To maintain similar channel counts in each level of the equiviariant networks, the total number of features actually decreases. For instance,
15 channels cannot use 15 irrep-1 features, as that yields $15 \times 3$ channels since irrep-1 is 3-dimensional. 

For a fair comparison, we preserved baseline channel budgets: 8 channels at the first encoder level in VM/PR++ and 16 in RDP, followed by multiples of 16 in deeper layers. Our equivariant design therefore respects these budgets: the first layer is restricted to irrep-0 and irrep-1 with a 1:1 ratio (since irrep-2 would exceed the small channel count), while deeper layers allow mixing in irrep-2, with a ratio of 5:2:1 for irrep-0, irrep-1 and irrep-2, resulting in an initial  $5\times1 + 2\times3 + 1\times5 = 16$ number of channels. 
Although RDP begins with 16 channels, we adopt the same principle for consistency.
We also experimented with other ratio combinations to investigate sensitivity to feature ratios (details in experiments section). 

For RDP, the equivariant encoder leads to a larger parameter reduction than in the other baselines, due to its substantially higher original capacity. To assess whether this reduction limits performance, we also evaluate a higher-capacity variant (Equi RDP dense) that replaces each convolution block in the encoder with two equivariant blocks.

\section{Experimental Setup}
\label{sec:exp}
\textbf{3.1. Datasets:}
We conducted our experiments on three main 3D brain MRI datasets: \textit{OASIS} \cite{marcus2007open} with 35 anatomical labels.
\textit{LPBA40} \cite{shattuck2008construction} with 54 anatomical labels.
\textit{MindBoggle} \cite{klein2017mindboggling} with 97 anatomical labels, with both cortical (62 labels) and subcortical (35 labels) regions.

\noindent
\textbf{3.2. Implementation Details:}
We implemented models in PyTorch using the escnn library \cite{cesa2022a} for steerable kernels.
Training hyperparameters (epochs, learning rate, pyramid levels) were kept identical between equivariant models and their baselines for fair comparison.

\noindent
\textbf{3.3. Evaluation Metrics:}
We employ \textit{Dice Similarity Coefficient (DSC)} for volumetric overlap, and {\textit{Average Symmetric Surface Distance (ASSD)} for surface alignment.

\section{Results and Discussion}
\label{sec:results}

\begin{table}[t]
  \centering
  \scriptsize
  \caption{Evaluation results on Brain MRI datasets: MindBoggle (97 ROIs), LPBA (54 ROIs), and OASIS (35 ROIs).}
  \label{tab:brain}
  \setlength{\tabcolsep}{2pt} 
  \renewcommand{\arraystretch}{1.05}

  \begin{adjustbox}{width=\columnwidth,center}
  \begin{tabular}{@{}l|cc|cc|cc@{}} 
    \toprule
    & \multicolumn{2}{c|}{\textbf{MindBoggle}}
    & \multicolumn{2}{c|}{\textbf{LPBA}}
    & \multicolumn{2}{c}{\textbf{OASIS}} \\
    \cmidrule(lr){2-3}\cmidrule(lr){4-5}\cmidrule(lr){6-7}
    \textbf{Method}
      & \textbf{DSC $\uparrow$} & \textbf{ASSD $\downarrow$}
      & \textbf{DSC $\uparrow$} & \textbf{ASSD $\downarrow$}
      & \textbf{DSC $\uparrow$} & \textbf{ASSD $\downarrow$} \\
    \midrule
    VM
      & 0.478 $\pm$ 0.101 & 2.217 $\pm$ 1.07
      & 0.629 $\pm$ 0.033 & 2.180 $\pm$ 0.25
      & 0.790 $\pm$ 0.028 & 0.725 $\pm$ 0.11 \\
    Equi VM
      & \textbf{0.494 $\pm$ 0.108} & \textbf{2.176 $\pm$ 1.09}
      & \textbf{0.633 $\pm$ 0.034} & \textbf{2.140 $\pm$ 0.24}
      & \textbf{0.793 $\pm$ 0.027} & \textbf{0.717 $\pm$ 0.11} \\
    \midrule
    Dual PR++
      & 0.535 $\pm$ 0.112 & 2.094 $\pm$ 1.11
      & 0.684 $\pm$ 0.024 & 1.864 $\pm$ 0.19
      & 0.817 $\pm$ 0.022 & 0.630 $\pm$ 0.08 \\
    Equi Dual \\PR++
      & \textbf{0.564 $\pm$ 0.107} & \textbf{1.928 $\pm$ 1.03}
      & \textbf{0.690 $\pm$ 0.023} & \textbf{1.834 $\pm$ 0.19}
      & \textbf{0.820 $\pm$ 0.020} & \textbf{0.616 $\pm$ 0.08} \\
    \midrule
    RDP Net
      & 0.607 $\pm$ 0.076 & 1.632 $\pm$ 0.68
      & 0.727 $\pm$ 0.018 & 1.610 $\pm$ 0.13
      & 0.831 $\pm$ 0.023 & 0.577 $\pm$ 0.08 \\
    Equi RDP \\Net
      & 0.608 $\pm$ 0.077 & 1.636 $\pm$ 0.70
      & 0.729 $\pm$ 0.018 & 1.599 $\pm$ 0.14
      & \textbf{0.832 $\pm$ 0.023} & \textbf{0.574 $\pm$ 0.09} \\
    Equi RDP \\dense
      & \textbf{0.611 $\pm$ 0.073} & \textbf{1.605 $\pm$ 0.66}
      & \textbf{0.730 $\pm$ 0.017} & \textbf{1.586 $\pm$ 0.13}
      & \textbf{0.832 $\pm$ 0.024} & 0.577 $\pm$ 0.093 \\
    \bottomrule
  \end{tabular}
  \end{adjustbox}

  \vspace{-0.5\baselineskip}
\end{table}

\begin{figure}[t]
  \centering
  \includegraphics[width=0.8\linewidth]{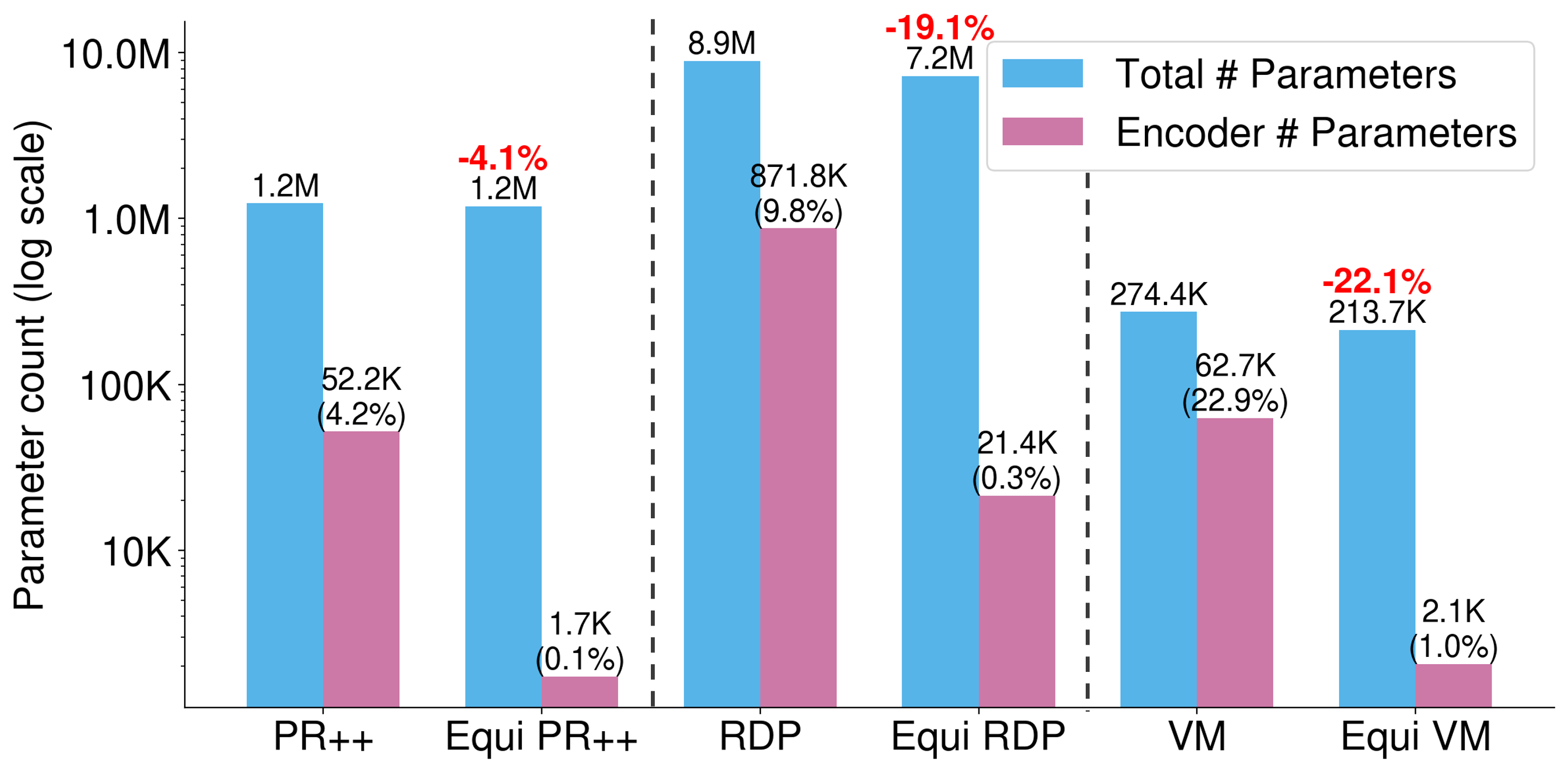}
  \caption{Parameter comparison (log scale) between baselines and equivariant counterparts. Red values show parameter reduction. Equivariant models significantly reduce parameters. }
  \label{fig:param}
  \vspace{-1.2\baselineskip}
\end{figure}

\begin{figure}[t]
  \centering
  \includegraphics[width=.9\linewidth]{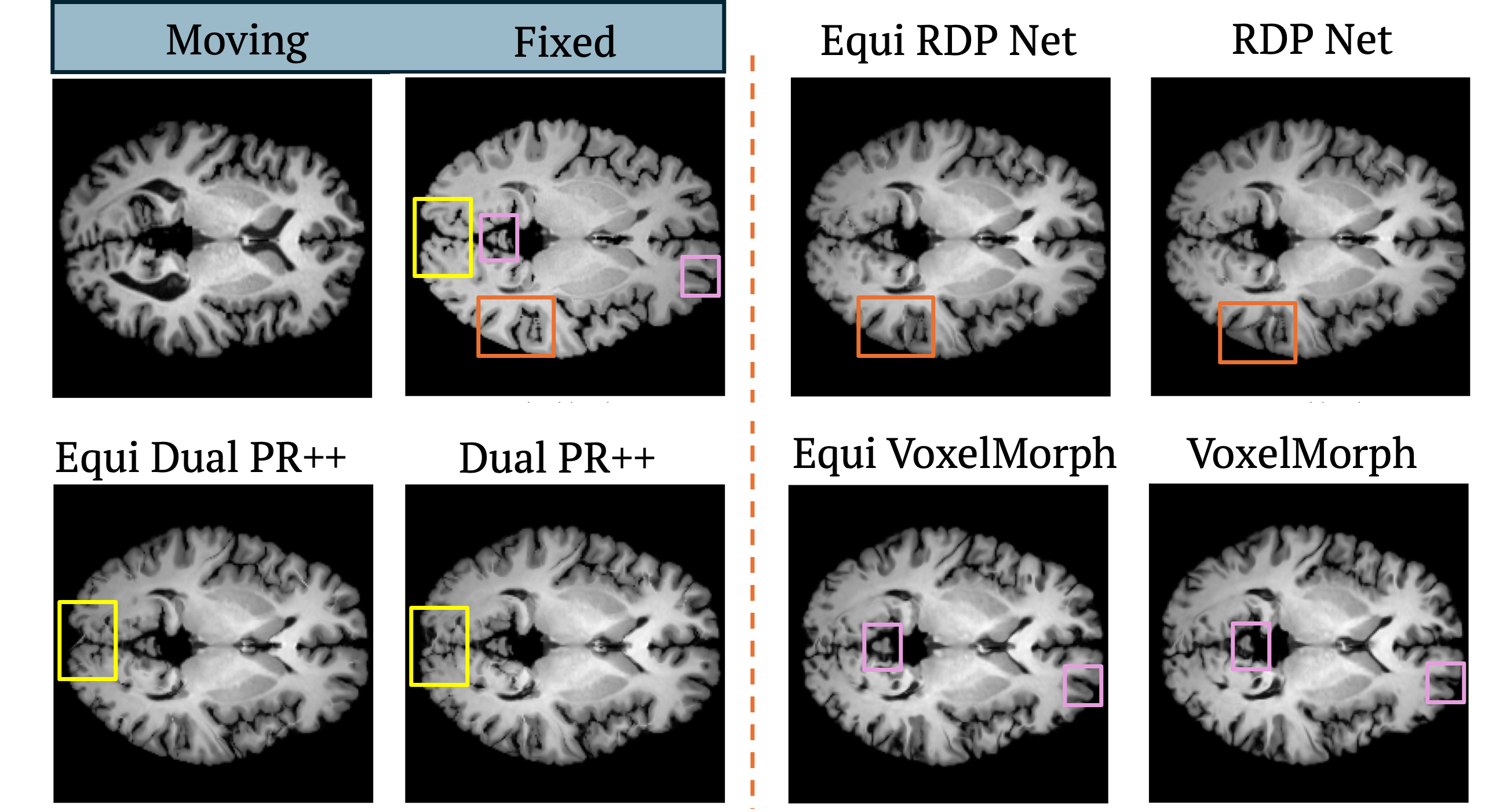}
  \caption{Qualitative results of MindBoggle dataset on different baselines and their equivariant version.}
  \label{fig:qual}
  \vspace{-10pt}
\end{figure}

\begin{table}[t]
  \centering
  \caption{Dice scores for rotated inputs. Equivariant models achieve higher average scores and lower standard deviation.}
  \label{tab:rot_once_dice}
  \setlength{\tabcolsep}{20pt}
  \renewcommand{\arraystretch}{0.95}

  \resizebox{\columnwidth}{!}{%
  \begin{tabular}{l|cccc}
    \toprule
    \textbf{Model} & \textbf{Avg} & \textbf{Std} & \textbf{Min} & \textbf{Max} \\
    \midrule
    Dual PR ++         & 0.554 & 0.107 & 0.416 & 0.684 \\
    Equi Dual PR ++    & 0.563 & 0.106 & 0.426 & 0.690 \\
    \midrule
    RDP           & 0.611 & 0.107 & 0.470 & 0.727 \\
    Equi RDP      & 0.628 & 0.098 & 0.491 & 0.729 \\
    \bottomrule
  \end{tabular}%
  }
  \vspace{-0.5\baselineskip}
\end{table}

\begin{figure}[t]
  \centering
  \includegraphics[width=0.9\linewidth]{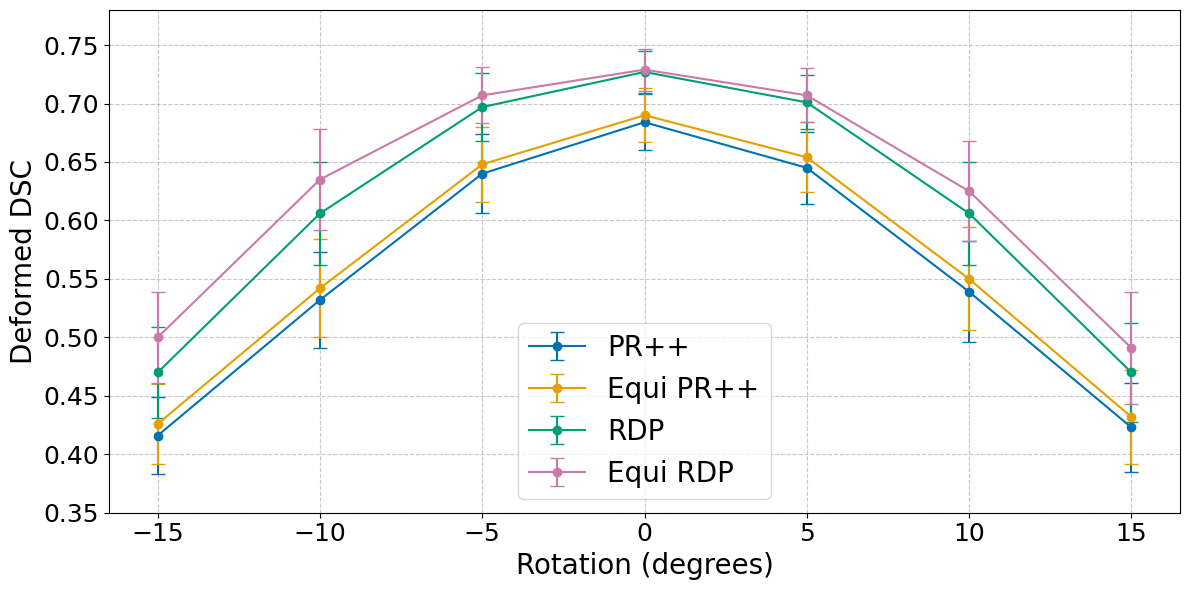}
  \caption{
  Dice scores for rotation of moving image only.}
  \label{fig:rot}
  \vspace{-0.5\baselineskip}
\end{figure}

  
  

\textbf{4.1. Parameter efficiency:}
Fig. \ref{fig:param} compares parameter counts between baseline and equivariant models. Equivariant versions contain 78\% (VM), 96\% (Dual-PR++), and 81\% (RDP) of baseline parameters, with reductions solely from encoder modifications. This parameter efficiency contributes to better generalizability and sample efficiency (which will be further discussed).

\noindent
\textbf{4.2. Evaluation results:}
 Table \ref{tab:brain} compares different baseline models with their corresponding equivariant versions on brain MRI datasets with symmetries.
As shown in the table, despite the fewer parameters, equivariant models demonstrate improved or comparable performance relative to their baseline counterparts. 
Qualitative results are provided in Fig.\ref{fig:qual}.


Not all models and datasets benefit equally from equivariant architectures.VM and Dual PR++ show greater improvement on MindBoggle compared to LPBA, and LPBA more than OASIS.
This pattern aligns with the fact that MindBoggle has more labeled anatomies (97 labels including 62 cortical labels), which provides a more suitable environment for equivariant models to excel.
 OASIS's coarse labeling reduces the importance of rotational symmetries, yielding only marginal improvements.
 
Model structure also influence the degree of performance gain. Most improvements are associated with the Dual-PR++ model, and can be attributed to the synergy between Dual-PR++'s correlation-based correspondence mechanism and rotation-equivariant features, which enhances the reliability of voxel correspondence estimation.

However, improvements on RDP-Net are marginal. RDP-Net is already a strong, high-capacity model with substantially more parameters than the other baselines (Fig. \ref{fig:param}). Replacing its encoder with an equivariant one sharply reduces parameters and adds constraints that limit flexibility and, in turn, the headroom to improve over the baseline. Overall, these results indicate that \textbf{the benefits of equivariance depend on both the dataset characteristics and the model’s capacity and architecture.}
We also note that the denser Equi RDP variant, which restores a small portion of the parameters removed by the equivariant encoder (contains 82\% of baseline parameter), achieves better performance than Equi-RDP and RDP, showing that even modest additional capacity allows the equivariant encoder to deliver greater improvements.

\noindent
\textbf{Statistical analysis:} For each model–dataset pair, we performed \textbf{paired Wilcoxon signed-rank tests} on per-case Dice scores. Overall, Eeuivariant variants achieved statistically significant improvements over their baselines ($p \leq 0.05$), with consistent gains, supporting the utility of the equivariant encoder when symmetry is present.



\noindent
\textbf{4.3. Results on rotated input:}
To test rotation equivariance, we conducted controlled experiments on LPBA test images. All models were trained on standard data; only test images were rotated. While perfect equivariance isn't expected due to non-equivariant decoders, we anticipate improved robustness from equivariant feature extractors.
Our evaluation consists of single-image rotation where one image was fixed and the other was rotated (0°, $\pm$5°, $\pm$10°, $\pm$15°) to simulate clinical scenarios with positioning variations, and restricted to dual-stream models.
As shown in Table \ref{tab:rot_once_dice} equivariant models consistently outperform baselines in DSC. While all models degrade with larger rotations, equivariant models maintain better performance, particularly Equi RDP (Fig.\ref{fig:rot}).




\begin{table}[t]
  \centering
  \tiny
  \caption{Equivariant channel configurations.}
  \label{tab:equi_channels}
  \setlength{\tabcolsep}{1pt}
  \renewcommand{\arraystretch}{0.95}

  \resizebox{\linewidth}{!}{%
  \begin{tabular}{l|ccc|c}
    \toprule
    \textbf{Model config} & \textbf{irrep-0} & \textbf{irrep-1} & \textbf{irrep-2} & \textbf{Total channels} \\
    \midrule
    Equi irreps 0            & 16 & 0 & 0 & $16\times 1 = 16$ \\
    Equi irreps 1            & 0  & 5 & 0 & $5\times 3 = 15$  \\
    Equi irreps 2            & 0  & 0 & 3 & $3\times 5 = 15$  \\
    \midrule
    Equi irreps 01 (4:4)     & 4  & 4 & 0 & $4\times1 + 4\times 3 = 16$ \\
    Equi irreps 01 (7:3)     & 7  & 3 & 0 & $7\times 1 + 3\times 3 = 16$ \\
    \midrule
    Equi irreps 012 (5:2:1)  & 5  & 2 & 1 & $5\times 1 + 2\times 3 + 1\times 5 = 16$ \\
    Equi irreps 012 (2:2:2)  & 2  & 2 & 2 & $2\times 1 + 2\times 3 + 2\times 5 = 18$ \\
    \bottomrule
  \end{tabular}%
  }\label{tab:ch}
  \vspace{-1.3\baselineskip}
\end{table}

\begin{figure}[t]
  \centering
  \includegraphics[width=0.9\linewidth]{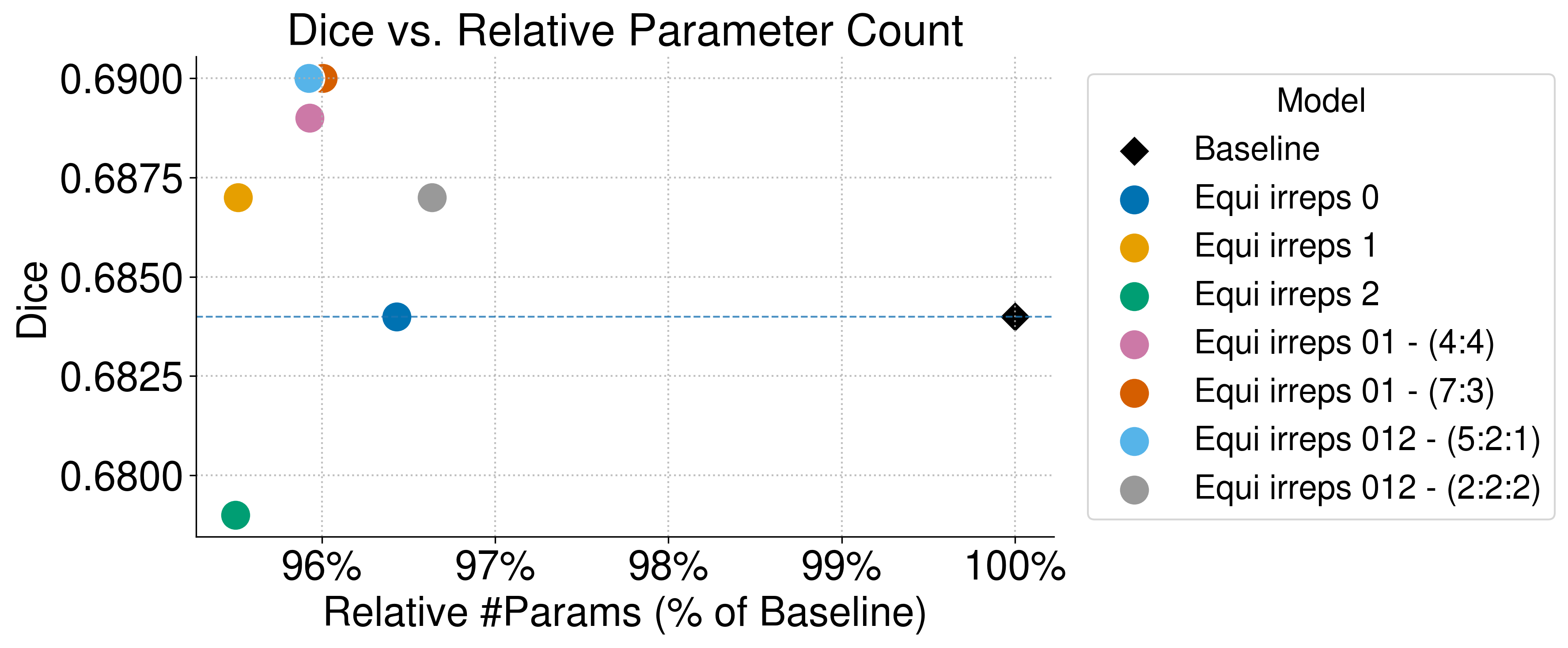}
  \caption{Comparison of performance of Equivarint models with different choices of irreps.}
  \label{fig:dice_vs_params}
  \vspace{-0.5\baselineskip}
\end{figure}



\begin{table}[t]
  \centering
  \caption{Evaluation results on the LPBA test set using different portions of the LPBA training set.}
  \label{tab:lpba_small}
  \setlength{\tabcolsep}{3pt}
  \renewcommand{\arraystretch}{0.95}

  \resizebox{\columnwidth}{!}{%
  \tiny
  \begin{tabular}{l|cc|cc}
    \toprule
    & \multicolumn{2}{c|}{\textbf{PR ++}}
    & \multicolumn{2}{c}{\textbf{Equi PR ++}} \\
    \cmidrule(lr){2-3}\cmidrule(l){4-5}
    \textbf{Training Portion}
      & \textbf{DSC} & \textbf{ASSD}
      & \textbf{DSC} & \textbf{ASSD} \\
    \midrule
    \textbf{LPBA (full)}
      & 0.684 $\pm$ 0.024 & 1.864 $\pm$ 0.195
      & \textbf{0.690} $\pm$ \textbf{0.023} & \textbf{1.834} $\pm$ \textbf{0.193} \\
    \textbf{LPBA (1/2)}
      & 0.675 $\pm$ 0.025 & 1.911 $\pm$ 0.199
      & \textbf{0.680} $\pm$ \textbf{0.024} & \textbf{1.883} $\pm$ \textbf{0.193} \\
    \textbf{LPBA (1/4)}
      & 0.665 $\pm$ 0.027 & 1.962 $\pm$ 0.210
      & \textbf{0.673} $\pm$ \textbf{0.025} & \textbf{1.916} $\pm$ \textbf{0.191} \\
    \textbf{LPBA (1/8)}
      & 0.653 $\pm$ 0.031 & 2.032 $\pm$ 0.233
      & \textbf{0.660} $\pm$ \textbf{0.028} & \textbf{1.987} $\pm$ \textbf{0.209} \\
    \bottomrule
  \end{tabular}%
  }
  \vspace{-1.2\baselineskip}
\end{table}

\noindent
\textbf{4.4. Effect of irreps ratio:}
\label{sec:irreps}
We investigated different ratios for scalar, vector, and rank-2 tensor features (irreps 0, 1, and 2) in the encoder design of the Dual-PR++ model to examine the effect of this hyperparameter on model performance on LPBA dataset.
For the mixed configurations (7:3, 4:4, 5:2:1, 2:2:2), we fixed the first layer to a 2:2 split of irrep-0 and irrep-1 following the design principle in \ref{sec:enc_design}. Target ratios were applied at deeper layers (16+ channels) to maintain similar channel counts (Table \ref{tab:ch}).
As illustrated in Fig. \ref{fig:dice_vs_params}, all combinations improve the parameter efficiency  over the baseline.
Using irrep-2 alone degrades performance, while other combinations achieve similar or better results. The best configuration achieving the highest Dice score with lowest parameter count is the (5:2:1) ratio for scalar, vector, and rank-2 tensor features, respectively.



\noindent
\textbf{4.5. Architectural variants:}
We also tested a fully equivariant VoxelMorph (encoder and decoder), but it underperformed both the encoder-only variant and baseline, likely due to the decoder's severely reduced capacity (~10k parameters, 3.5\% of baseline) and constraints limiting flexibility for complex deformations. This confirms registration benefits from equivariant encoders but requires high-capacity decoders.

\noindent
\textbf{4.6. Effect of dataset size:}
\label{sec:dataset_size}
To investigate training set size effects, we trained Dual-PR++ and equivariant Dual-PR++ on different portions of the LPBA training set while keeping the test set fixed. As presented in Table \ref{tab:lpba_small},
both models degrade with less data, but the equivariant model consistently outperforms the baseline with larger gaps on smaller datasets. Notably, the equivariant model on small datasets matches baseline performance on larger datasets, confirming that rotational symmetry inductive bias improves sample efficiency.

\noindent
\textbf{4.7. Limitations:}
Equivariant constraints reduce learnable parameters, which generally improves efficiency but can limit flexibility when reductions are too restrictive, as seen in RDP where equivariance only maintains baseline performance on some datasets, and a denser equivariant version improved over both.

Steerable kernels introduce computational overhead during training due to equivariance constraints, though they can be exported as standard CNNs for inference with similar performance.

\section{Conclusion}

In this work, we demonstrated that integrating rotation-equivariant SE(3) encoders into deformable brain MRI registration networks improves performance and parameter efficiency. We evaluated this approach by replacing standard encoders in three baseline architectures (VoxelMorph, Dual-PRNet++, and RDP Net) across multiple brain MRI datasets.

Our experiments yield three key findings. First, equivariant models achieve comparable or improved registration accuracy (Dice/ASSD) while using only 78-96\% of the baseline parameters. Second, experiments on rotated inputs confirm their superior robustness to orientation variations, a crucial advantage given clinical positioning variability. Third, the models exhibit greater sample efficiency, achieving strong performance with reduced training data.

This work provides the first systematic investigation of SE(3) equivariance in deformable medical image registration. Our findings confirm that incorporating geometric inductive biases is a critical step toward building more robust, efficient, and accurate registration models. Future work could explore adaptive mechanisms for leveraging these biases and applications to other anatomical structures.

\section{Compliance with Ethical Standards}
\label{sec:ethics}
This research study was conducted retrospectively using human subject data made available in open access by Oasis \cite{marcus2007open}, LPBA40 \cite{shattuck2008construction} and MindBoggle \cite{klein2017mindboggling}. Ethical approval was not required as confirmed by the licenses attached with these open access datasets.
\section{Acknowledgments}
\label{sec:acknowledgments}
The authors report no conflicts of interest.

\bibliographystyle{IEEEbib}
\bibliography{strings,refs}

\end{document}